\documentclass[runningheads]{llncs}

\newcommand{\stddev}[1]{\textcolor{gray!50}{#1}} 
\usepackage[T1]{fontenc}
\usepackage[dvipsnames]{xcolor}

\usepackage{graphicx}
\usepackage{subcaption}
\usepackage{hhline}
\usepackage{hyperref}
\usepackage{color}

\usepackage[misc]{ifsym}

\makeatletter
\renewcommand\@fnsymbol[1]{} 
\makeatother

\usepackage{booktabs}
\usepackage{array}
\usepackage{tabularx}
\usepackage{float}
\usepackage{multirow} 
\usepackage{siunitx} 
\usepackage[export]{adjustbox}
\usepackage{graphicx}

\usepackage{listings}
\usepackage{tcolorbox}

\begin{document}
%
\title{Can LLM-Generated Textual Explanations Enhance Model Classification Performance? An Empirical Study}

\titlerunning{ } 

%


\author{Mahdi Dhaini \textsuperscript{\Letter}
 \and
Juraj Vladika \and
Ege Erdogan \and Zineb Attaoui \and Gjergji Kasneci
}

%
\authorrunning{ }
%
\institute{Technical University of Munich, School of Computation, Information and Technology, Department of Computer Science, Munich, Germany
\\
\email{firstname.lastname@tum.de} \thanks{\Letter\ Corresponding author} 
}


\maketitle              
%


\begin{abstract}
In the rapidly evolving field of Explainable Natural Language Processing (NLP), textual explanations, i.e., human-like rationales, are pivotal for explaining model predictions and enriching datasets with interpretable labels. Traditional approaches rely on human annotation, which is costly, labor-intensive, and impedes scalability. In this work, we present an automated framework that leverages multiple state-of-the-art large language models (LLMs) to generate high-quality textual explanations. We rigorously assess the quality of these LLM-generated explanations using a comprehensive suite of Natural Language Generation (NLG) metrics.
Furthermore, we investigate the downstream impact of these explanations on the performance of pre-trained language models (PLMs) and LLMs across natural language inference tasks on two diverse benchmark datasets. Our experiments demonstrate that automated explanations
exhibit highly competitive effectiveness compared to human-annotated explanations in improving model performance. Our findings underscore a promising avenue for scalable,
automated LLM-based textual explanation generation for extending NLP datasets and enhancing model performance.

\keywords{Explainable natural language processing \and Natural language explanations \and Natural language inference \and Large language models.}
\end{abstract}
\section{Introduction}
Recent NLP advancements are driven by PLMs and LLMs, achieving state-of-the-art results across various tasks \cite{brown2020language}. However, their black-box nature limits understanding of their predictions, prompting increased interest in Explainable NLP, where methods from Explainable AI explain model decision-making \cite{soegaard-2022-xnlp-book} to enhance trust and transparency, which is essential for advancing practical applications in sensitive domains.

A key challenge in Explainable NLP is the lack of definitive ground-truth explanations \cite{lei2016rationalizing}. Researchers address this by collecting human-generated textual explanations, creating \textit{explainable datasets} \cite{deyoung2020eraser}. These datasets serve both as benchmarks for evaluating model-generated explanations and as training data to improve models’ predictive performance \cite{wiegreffe2020annotated}. 
However, human annotation is resource-intensive, impacting dataset scale and quality \cite{rajani2019explain}. Recently, leveraging LLMs' text-generation capabilities for explanations has gained attention \cite{wei2022chain}, though evaluating these explanations' quality and effectiveness in downstream tasks remains an open research question.

In this paper, we address these critical gaps by focusing on two primary objectives. First, we leverage multiple LLMs to automatically generate textual explanations and rigorously evaluate their quality using a comprehensive suite of metrics. Second, we investigate how the incorporation of these LLM-generated explanations impacts the performance of various PLMs and LLMs on downstream tasks, particularly within the NLI framework. 

Our work is guided by the following research question: 
\textit{How do LLM-generated textual explanations impact the performance of PLMs and LLMs on downstream predictive tasks?}
Our contributions are as follows:
\begin{itemize}
    \item We employ four LLMs of varying sizes and complexity to automatically generate explanations for two explainable NLI datasets in both zero-shot and few-shot settings.
    \item We evaluate the quality of the generated explanations using multiple metrics, including both reference-based measures and an innovative LLM-based evaluation approach.
    \item We examine the impact of incorporating LLM-generated explanations during both fine-tuning and inference, comparing their effects against human-annotated explanations and a no-explanation baseline across four distinct BERT-based models and three LLMs.\footnote{We release our code on \href{https://github.com/dmah10/helpful-natural-language-explanations}
    {\textbf{GitHub}}}
\end{itemize}


\section{Background and Related Work}

\textbf{Natural Language Inference}
(NLI)
is one of the most fundamental NLP tasks \cite{gubelmann2024capturing}. The goal is, given two pieces of text, a premise and a hypothesis, to determine a logical relation between them as one of the three classes: entailment, contradiction, or neutral. 
The turning point in NLI was the construction of the Stanford NLI (SNLI) corpus in 2015 \cite{bowman-etal-2015-large}, a dataset of half a million examples, constructed with crowd-sourced effort where photos were captioned and then paired with entailed, contradicted, or neutral sentences written by annotators. 
Modern PLMs like BERT \cite{devlin2019bert} and RoBERTa \cite{liu2019roberta}, as well as autoregressive LLMs like GPT, can often solve popular NLI datasets with an above-human performance, owing to the linguistic patterns and world knowledge acquired during their pre-training on huge corpora.

%

\noindent\textbf{Explainable NLP and Datasets}
The growing interest in Explainable NLP is evident from multiple surveys like 
\cite{madsen-xnlp-survey-2023,wiegreffe2020annotated},
some
addressing specific tasks
or
methods \cite{Mardaoui-2021-lime-survey}.
Interest in explainable NLP has led to the creation of explainable datasets for tasks such as hate-speech classification \cite{mathew2021hatexplain} and claim verification \cite{vladika-etal-2025-step}.
 A comprehensive review of these datasets is provided in \cite{wiegreffe2020annotated}. Textual explanations typically fall into highlights, structured, or free-text (natural language) categories and are annotated by authors, experts, and crowd-sourcing 
 with most datasets relying on human annotators. However, human annotation presents several challenges. Collecting high-quality explanations is time-consuming and resource-intensive \cite{hartmann-2022-survey-human-explanation-performance}. Human annotators' explanations may also suffer from subjectivity and inconsistency, potentially hindering model performance rather than aiding it \cite{yao-2023-human-explanations-helpful}. Additionally, the diversity in explanation types introduces further complexities \cite{tan2021diversity}.

\noindent \textbf{Generating LLM-explanations}
Due to the limitations of human-annotated explanations, recent research has explored using LLMs to generate Natural Language Explanations (NLE) and justifications for model decisions. Compared to traditional post-hoc feature attribution methods, NLEs provide human-readable justifications, which can enhance transparency and user understanding.
\cite{mishra-etal-2024-characterizing-rationalizers} employed LLMs as rationalizers for knowledge-intensive tasks such as multiple-choice question answering. 
\cite{wang-etal-2025-cross-refine} investigated improving LLM-generated NLE quality through a tandem learning setup. 
\cite{wei-jie-etal-2024-interpretable-reasoning-nle} examined how various prompting techniques, such as CoT, can improve NLEs on commonsense reasoning tasks. 
These studies demonstrate the growing interest in leveraging LLMs to generate and refine NLEs, particularly for tasks requiring explanation-driven reasoning. However, none of the previously mentioned works investigate how extending datasets with LLM-generated explanations can impact the performance of PLMs and LLMs on downstream tasks.     

\noindent \textbf{Evaluating NLEs} \label{sec:background_evaluation}
NLEs are text snippets and can be evaluated with standard NLG metrics \cite{schmidtova-etal-2024-automatic-metrics}. When human-written (gold) references exist, \textit{reference-based} metrics are applicable. Traditional metrics, such as BLEU \cite{papineni-etal-2002-bleu} and ROUGE \cite{lin-2004-rouge}, assess word overlaps between generated and reference texts. However, these metrics have become less suitable with the rise of LLMs, as they penalize expressive variations in wording. Consequently, semantic metrics like the embedding-based BERTScore \cite{Zhang2020BERTScore} and distribution-based MAUVE \cite{pillutla2021mauve} have gained popularity. Recently, evaluation methods using \textit{LLM-as-judge} metrics have emerged, employing crafted prompts for LLMs to return numerical scores assessing generated texts, exemplified by G-Eval \cite{liu-etal-2023-G-eval}.

\noindent \textbf{Closely related work}
Among the previously mentioned works, the closest to ours is \cite{yao-2023-human-explanations-helpful}, which investigates how human explanations can impact the predictions of two PLMs. However, their study is limited to BART and T5 and focuses solely on human explanations. Additionally, \cite{hartmann-2022-survey-human-explanation-performance} reviews studies employing different types of human explanations (highlights, structured, and free-text) to improve NLP models. However, they solely review studies incorporating human-annotated explanations.
In contrast, while we also incorporate human explanations, our primary focus is on generating and investigating LLM-generated NLEs. We evaluate the impact of these explanations on four PLMs, including the recent ModernBERT \cite{warner2024smarter}, as well as three LLMs of varying sizes. 
\section{Experimental Setup}
We designed a comprehensive experimental framework that systematically integrates both human- and LLM-generated explanations into two benchmark datasets. 
For reproducibility,  we provide all prompt templates for explanation generation, evaluation, and LLM performance evaluation in the repository.

\subsection{Datasets}

We use two datasets in our experiments. The first dataset, e-SNLI \cite{camburu2018e}, is an extension of the SNLI dataset with human-annotated natural language explanations. It contains premise-hypothesis pairs labeled as entailment, neutral, or contradiction, depending on how the premise relates to the hypothesis. The second dataset is the HealthFC dataset \cite{vladika-etal-2024-healthfc}. It consists of 750 health-related claims, labeled by medical experts and backed with evidence from systematic reviews and clinical trials. Each claim is paired with pieces of evidence and includes a verdict (supported, refuted, not enough information), as well as brief explanations for the verdict. For our experiments, we extracted a balanced subset of e-SNLI consisting of 840 examples, ensuring an equal representation of entailment, neutral, and contradiction instances. This subset size was deliberately chosen to closely match the 750-instance HealthFC dataset, enabling a fair and controlled comparison across our evaluation framework.
Even though HealthFC is officially a dataset for automated fact-checking (claim verification), it is common to model this task as an NLI task. We provide further details on the datasets with examples in Appendix \ref{app:datasets}.

\subsection{Generating Natural Language Explanations with LLMs}
In our pipeline, we focus on generating NLEs using multiple LLMs.
We further extend both datasets we consider with explanations we generate using GPT-4o mini \cite{hurst2024gpt}, 
Mixtral-7B \cite{jiang2024mixtral}, 
Gemma2-9B \cite{team2024gemma}, 
and LLama3-70B \cite{dubey2024llama}.
For Mixtral-7B, Gemma2-9B, and LLama3-70B we use the APIs provided by Groq\footnote{\url{https://groq.com/}} 
while for GPT-4o mini we use OpenAI APIs\footnote{\url{https://platform.openai.com/}}.
We selected LLMs ranging in size from 7B to 70B parameters\footnote{along with GPT-4o mini, whose exact size is unknown.}, to analyze how these factors influence both the quality of the generated text and the impact of generated explanations on downstream task performance. The rationale for selecting diverse LLMs, rather than models within the same family differing only in size, is to ensure a broader variety in the \textit{sources of explanations}. We discuss later in the paper how this approach could be expanded in future work.

We generate explanations from the four LLMs under two settings: few-shot and zero-shot. After initial prompt validation, we explicitly instructed LLMs not to reveal or hint at labels in their explanations to avoid biasing the evaluation during inference. The few-shot setting examines if LLM explanations improve after exposure to human-written examples and evaluates the impact of these explanations on downstream tasks. Both zero-shot and few-shot prompts are provided in our repository; the few-shot prompts include four (\textit{premise-hypothesis-explanation}) examples from the dataset. {Due to our hardware constraints, we do not perform any
memory-heavy
approaches 
like fine-tuning of LLMs or reinforcement learning. We leave these for future work. We provide more details on the explanations generation process, including prompts used in Appendix \ref{app:nle-generation}.}

\subsection{Evaluating LLM-Natural Language Explanations} \label{sec:evaluation_metrics}
As we focus on generating natural language explanations, we evaluate their quality using some of the widely adopted metrics in NLG research we described in Section~\ref{sec:background_evaluation}. 
We compare LLM-generated explanations with human-provided explanations from our selected datasets.
Specifically, we employ the widely used BLEU, ROUGE, and BERTScore metrics. 
Beyond these conventional metrics, we incorporate the recent MAUVE, a distribution-based metric that quantifies the divergence between generated and human-written texts using Kullback–Leibler (KL) divergences in a quantized embedding space and also the \textit{LLM-as-judge} G-Eval framework that has been increasingly used in recent NLG research. 
We use G-Eval to measure human likeness in LLM-generated explanations, in particular, the clarity, coherence, and structure of the LLM-generated explanation. We refer readers to Appendix \ref{app:evaluation} for further details on G-Eval computation, implementation specifics, and metric libraries.

\subsection{Models for NLI Predictions}
\textbf{Fine-tuning PLMs}.
For the downstream NLI task predictions, we use four PLMs (BERT 
, DeBERTa \cite{he2020deberta}
, RoBERTa 
, and ModernBERT \cite{warner2024smarter}). 
For each run with a certain kind of or without explanations, we perform a 80/20 train/test split and fine-tune the PLMs on the train set for 10 epochs using the AdamW optimizer with a learning rate of 3e-6 for ModernBERT and 1e-5 for the other PLMs. We repeat this five times with a stratified 5-fold cross-validation and report results averaged over the five splits. 

\noindent\textbf{Experiments with LLMs.}
We also use three LLMs: GPT-4o mini
Qwen 2.5 (7B) and Llama3.3-70B. 
For GPT, we use the OpenAI API, and for the two open-source LLMs, we use the API provided by Together AI 
. We give the LLM the premise-hypothesis (or claim-evidence) pairs as input, and optionally add the human- or LLM-generated explanations at the end of the hypothesis for e-SNLI and the claim for HealthFC.
We adopt a zero-shot inference approach without fine-tuning. Instead, the generated explanations are directly appended to the hypothesis in the prompt. 
Zero-shot inference is well established in current literature as a resource-efficient method that leverages the inherent generalization capabilities of LLMs without additional overhead. Moreover, we do not adopt resource-intensive approaches such as fine-tuning for LLMs, even with the existence of lighter approaches like PEFT, as the primary focus of this study is to measure the impact of different explanations on the performance, rather than to compare zero-shot with fine-tuned LLMs performance. We provide further details on prompting the LLMs in Appendix \ref{app:models}.
{Our experimental setup covers the complete cross-product of explanation methods and classification models, covering all possible combinations, including cases where identical LLMs function in both explainer and classifier roles.}

\section{Analysis and Discussion}

\begin{figure}[!]
  \centering
  \begin{subfigure}[b]{\textwidth} 
    \includegraphics[max width=\linewidth,
                 max height=0.8\textheight]{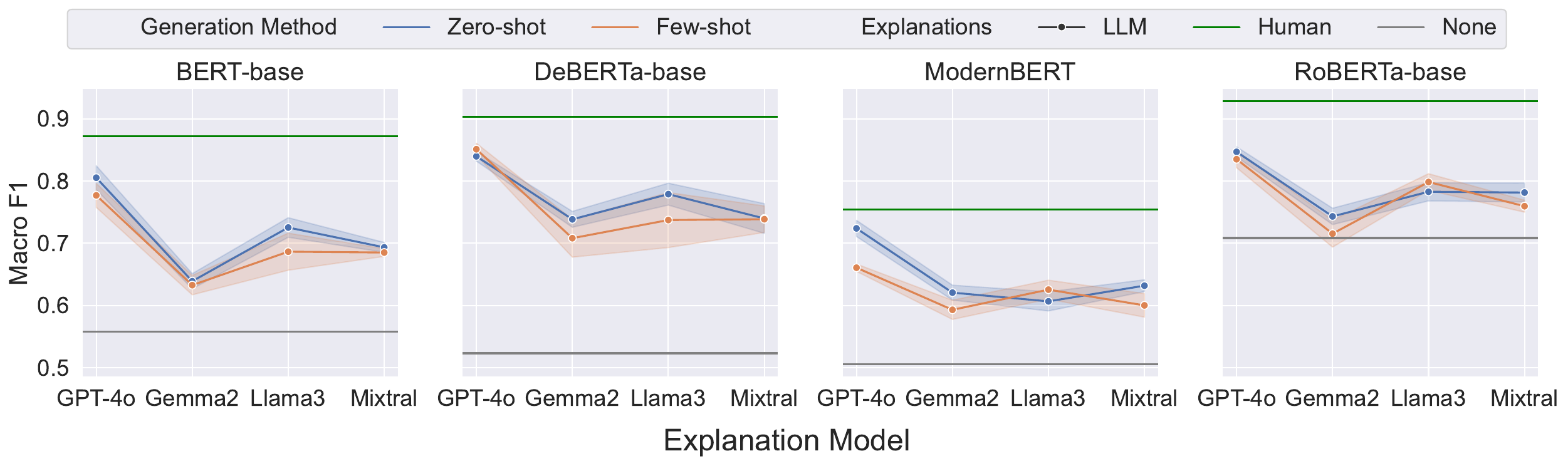}
    
    \caption{PLMs on e-SNLI}
    \label{fig:esnli_plms_macro}
  \end{subfigure}\hfill
  \begin{subfigure}[b]{\textwidth} 
    \includegraphics[max width=\linewidth,
                 max height=0.8\textheight]{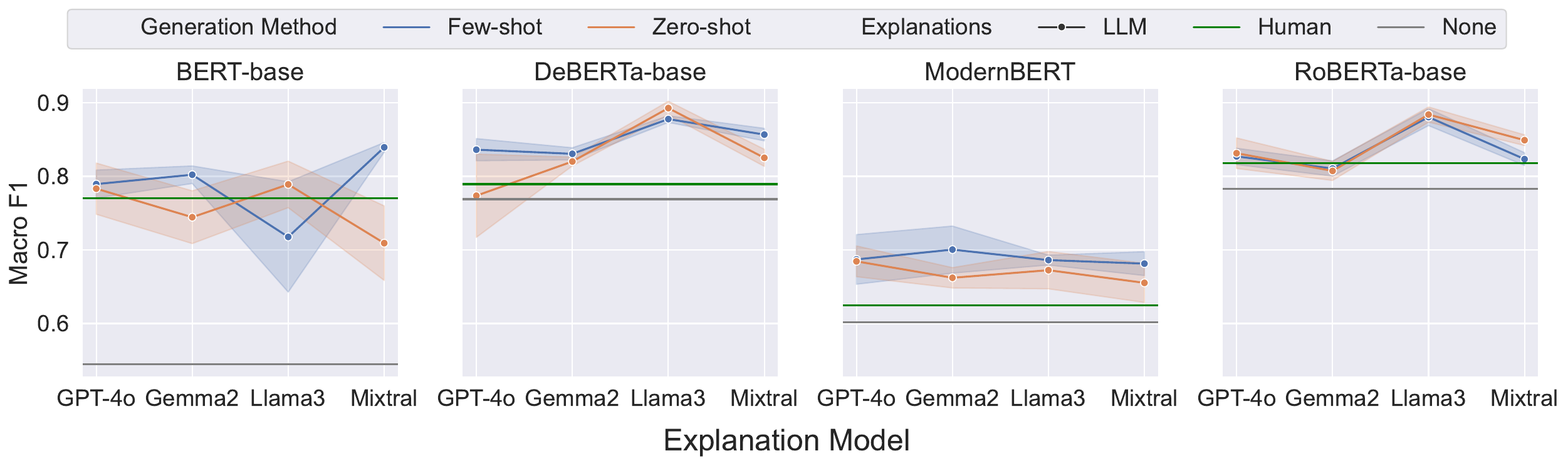}
    \caption{PLMs on HealthFC}
    \label{fig:hfc_plms_macro}
  \end{subfigure}

  \vspace{1em}

  \begin{subfigure}[b]{\textwidth}
    \includegraphics[max width=\linewidth,
                 max height=2\textheight]{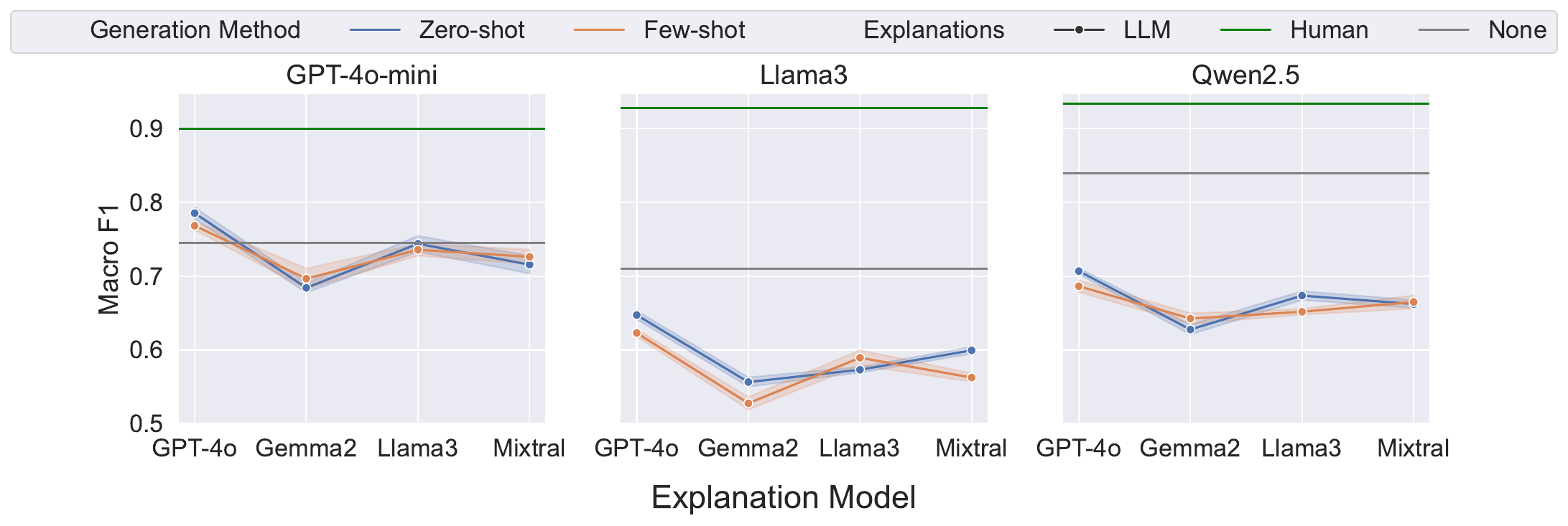}
    \caption{LLMs on e-SNLI}
    \label{fig:esnli_llms_macro}
  \end{subfigure}\hfill
  \begin{subfigure}[b]{\textwidth}
    \includegraphics[max width=\linewidth,
                 max height=0.9\textheight]{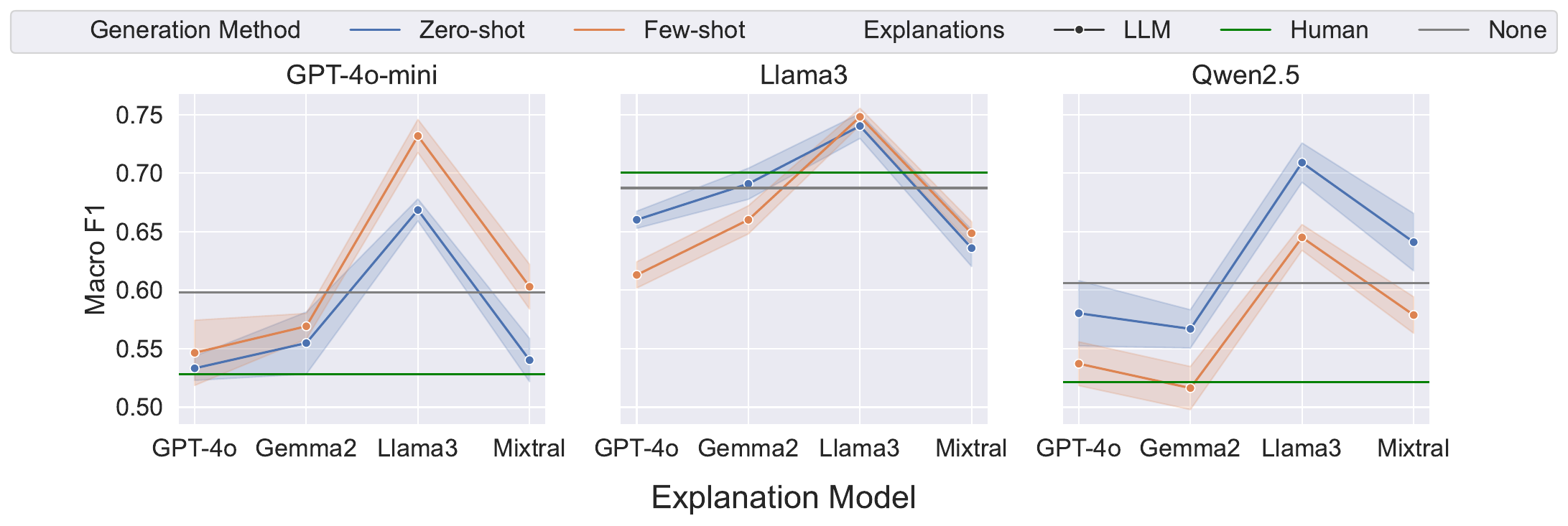}
    \caption{LLMs on HealthFC}
    \label{fig:hfc_llms_macro}
  \end{subfigure}

  \caption{%
    (Zoom in for better reading) Plots of the models’ performance on e-SNLI and HealthFC.  
    \textbf{Top row (a–b)} shows average \emph{Macro F1} for the four PLMs (BERT-base, DeBERTa-base, ModernBERT, RoBERTa-base);  
    \textbf{bottom row (c–d)} shows average \emph{Macro F1} for the three LLMs (GPT-4o mini, Llama3, Qwen2.5).  
    In each panel, bars are grouped by explanation input condition: no explanations (\textcolor{gray}{\textbf{gray}}), human explanations (\textcolor{ForestGreen}{\textbf{green}}), and explanations generated by four LLMs in zero-shot (\textcolor{blue}{\textbf{blue}}) vs.\ few-shot (\textcolor{orange}{\textbf{orange}}) settings.
  }
  \label{fig:combined-main-macro}
\end{figure}

Our results stem from an extensive experimental design covering multiple dimensions. Specifically, we evaluated two NLI datasets (e-SNLI and HealthFC), employed four different LLMs to generate explanations, and tested each in both zero-shot and few-shot settings, yielding 16 distinct explanation generation scenarios. We present the evaluation results in Table \ref{tab:explanation_metrics_evaluation_results}. 
Furthermore, we assessed downstream classification performance across four PLMs and three LLM classifiers. By analyzing metrics such as accuracy and macro F1 across these diverse combinations spread across Figure 
\ref{fig:combined-main-macro}
and Tables \ref{tab:plms_comparison}, \ref{tab:llms_comparison}, our study offers a comprehensive insight into how various explanation generation strategies affect NLI performance. While possible that our insights are specific only to the two chosen datasets, we try to make our takeaways general and widely applicable.



\subsection{Generation and Evaluation of LLM-explanations} \label{sec:eval_exp}

Table \ref{tab:explanation_metrics_evaluation_results} presents average metric scores for LLM-generated explanations. GPT-4o mini generally scores highest on e-SNLI, while Llama3-70B leads on HealthFC. GPT-4o mini outperforms others on e-SNLI in BLEU, ROUGE-1, and BERTScore-F1, whereas Llama3-70B excels in these metrics for HealthFC. GPT-4o mini consistently achieves top G-Eval scores, suggesting its explanations align closely with human judgment. However, G-Eval score differences across models are small, indicating similar overall quality. Mistral-7B achieves the highest MAUVE scores in multiple settings, implying greater diversity and coherence.
\begin{table}[!th]
    \caption{Average scores of the evaluation metrics across different LLMs on e-SNLI and HealthFC datasets in zero-shot and few-shot settings. The highest value for each metric is highlighted in \textbf{bold}.}
    \centering
    \setlength{\tabcolsep}{4pt}
    \renewcommand{\arraystretch}{1.3}
    \scriptsize
    \begin{tabularx}{\textwidth}{l l S[table-format=1.3] S[table-format=1.3] S[table-format=1.3] S[table-format=1.3]} 
        \toprule
        \textbf{Dataset} & \textbf{Metric} & \textbf{Gemma2-9B} & \textbf{Mistral-7B} & \textbf{Llama3-70B} & \textbf{GPT4o-mini} \\
        \midrule
        \multirow{5}{*}{\shortstack{e-SNLI \\ (zero-shot)}}
        & BLEU & 0.032 & 0.033 & 0.029 & \textbf{0.039} \\
        & ROUGE-1 & 0.295 & 0.314 & 0.277 & \textbf{0.333} \\
        & BERTScore F1 & 0.876 & 0.876 & 0.872 & \textbf{0.881} \\
        & MAUVE & 0.004 & \textbf{0.047} & 0.013 & 0.029 \\
        & G-Eval & 0.171 & 0.166 & 0.165 & \textbf{0.176} \\
        \midrule
        \multirow{5}{*}{\shortstack{e-SNLI \\ (few-shot)}}
        & BLEU & 0.037 & 0.043 & 0.037 & \textbf{0.051} \\
        & ROUGE-1 & 0.299 & 0.352 & 0.316 & \textbf{0.366} \\
        & BERTScore F1 & 0.878 & 0.882 & 0.878 & \textbf{0.885} \\
        & MAUVE & 0.004 & \textbf{0.107} & 0.040 & 0.084 \\
        & G-Eval & 0.163 & 0.170 & 0.160 & \textbf{0.174} \\
        \midrule
        \multirow{5}{*}{\shortstack{HealthFC \\ (zero-shot)}}
        & BLEU & 0.017 & 0.022 & \textbf{0.030} & 0.024 \\
        & ROUGE-1 & 0.269 & 0.285 & \textbf{0.313} & 0.292 \\
        & BERTScore F1 & 0.881 & 0.878 & \textbf{0.883} & 0.883 \\
        & MAUVE & 0.004 & \textbf{0.115} & 0.083 & 0.023 \\
        & G-Eval & 0.197 & 0.194 & 0.192 & \textbf{0.214} \\
        \midrule
        \multirow{5}{*}{\shortstack{HealthFC \\ (few-shot)}}
        & BLEU & 0.018 & 0.023 & \textbf{0.030} & 0.023 \\
        & ROUGE-1 & 0.261 & 0.294 & \textbf{0.309} & 0.291 \\
        & BERTScore F1 & 0.884 & 0.881 & \textbf{0.886} & 0.884 \\
        & MAUVE & 0.004 & 0.180 & \textbf{0.197} & 0.095 \\
        & G-Eval & 0.199 & 0.187 & 0.192 & \textbf{0.205} \\
        \bottomrule
    \end{tabularx}
    \label{tab:explanation_metrics_evaluation_results}
\end{table}  
Scores slightly improve from zero-shot to few-shot settings, particularly BLEU and ROUGE-1 on e-SNLI, but these improvements are minor, indicating limited benefit from in-context examples. Additionally, model size alone doesn't ensure better performance; smaller models like Gemma2-9B and Mistral-7B sometimes perform competitively or better.
Our analysis shows LLMs do not consistently prefer their own explanations. Human explanations generally provide more significant performance gains, especially on e-SNLI. GPT-4o mini excels on e-SNLI and Llama3-70B on HealthFC, with BLEU, ROUGE, and BERTScore strongly correlating with downstream improvements.

Overall, while scores improve slightly between zero-shot and few-shot settings (notably in BLEU and ROUGE-1 on e-SNLI), these improvements are marginal. This indicates that providing in-context examples from the dataset does not significantly enhance the generated explanations according to these metrics. Furthermore, model size alone does not guarantee better performance, as seen when comparing Gemma2-9B, Mistral-7B, and Llama3-70B, where smaller models sometimes achieve competitive or even higher scores.

\subsection{Influence of Explanations on the Performance of PLMs}

\textbf{Both human and LLM explanations improve PLMs' performance.} Results in
Figures \ref{fig:esnli_plms_macro}, \ref{fig:hfc_plms_macro} 
show that for both of our datasets, incorporating explanations generated both by humans and LLMs result in better predictive performance compared to the baseline of no explanations independent of the LLM used to generate the explanations. This could be related to the explanations providing additional information beneficial for the task, and the models learning to use that information since they are trained with the explanations as well. 

\noindent\textbf{Relative benefit from human and LLM explanations varies between datasets.} Table \ref{tab:plms_comparison} displays the change in performance after incorporating LLM-generated explanations, compared to human explanations and the no-explanation baseline. Most significantly, LLM-generated explanations lead to better performance than human explanations with the HealthFC dataset, but worse performance on e-SNLI. This indicates that even though LLM-generated explanations are consistently more beneficial than having no explanations, humans can write more beneficial explanations than LLMs on certain datasets. 

\begin{table}[t!]
\scriptsize
\small
    \centering
    \caption{\label{tab:plms_comparison}Performance \textbf{impact} by LLM-generated explanations over the baseline of no explanations and human-written explanations, \textbf{averaged over the four PLMs} we have used as classifiers. Subscripts indicate standard deviations.}

    \begin{subtable}{\textwidth}
        \centering
        \caption{e-SNLI}
        \begin{tabular}{llc|c|c|c}
        \hline
        & & \multicolumn{4}{c}{\textbf{Improvement over}}  \\
        & & \multicolumn{2}{c|}{\textbf{No Explanations}} & \multicolumn{2}{c}{\textbf{Human Explanations}} \\
        \hhline{~~----}
        \textbf{Explainer} & \textbf{Method} & \textbf{Accuracy} & \textbf{Macro F1} & \textbf{Accuracy} & \textbf{Macro F1} \\
        \hline
        \textbf{GPT-4o} & \textbf{Few-shot} & $0.171_{\stddev{0.087}}$ & $0.179_{\stddev{0.088}}$ & $-0.119_{\stddev{0.087}}$ & $-0.120_{\stddev{0.088}}$ \\
         & \textbf{Zero-shot} & $0.194_{\stddev{0.066}}$ & $0.204_{\stddev{0.066}}$ & $-0.096_{\stddev{0.066}}$ & $-0.095_{\stddev{0.066}}$ \\
        \hline
        \textbf{Gemma2} & \textbf{Few-shot} & $0.083_{\stddev{0.070}}$ & $0.091_{\stddev{0.068}}$ & $-0.207_{\stddev{0.070}}$ & $-0.208_{\stddev{0.068}}$ \\
         & \textbf{Zero-shot} & $0.087_{\stddev{0.066}}$ & $0.095_{\stddev{0.066}}$ & $-0.204_{\stddev{0.066}}$ & $-0.204_{\stddev{0.066}}$ \\
        \hline
        \textbf{Llama3} & \textbf{Few-shot} & $0.142_{\stddev{0.076}}$ & $0.151_{\stddev{0.076}}$ & $-0.148_{\stddev{0.076}}$ & $-0.148_{\stddev{0.076}}$ \\
         & \textbf{Zero-shot} & $0.113_{\stddev{0.088}}$ & $0.122_{\stddev{0.089}}$ & $-0.177_{\stddev{0.088}}$ & $-0.177_{\stddev{0.089}}$ \\
        \hline
        \textbf{Mixtral} & \textbf{Few-shot} & $0.110_{\stddev{0.072}}$ & $0.117_{\stddev{0.072}}$ & $-0.180_{\stddev{0.072}}$ & $-0.182_{\stddev{0.072}}$ \\
         & \textbf{Zero-shot} & $0.119_{\stddev{0.065}}$ & $0.128_{\stddev{0.065}}$ & $-0.171_{\stddev{0.065}}$ & $-0.171_{\stddev{0.065}}$ \\
         \hline
        \end{tabular}        
    \end{subtable}

    \begin{subtable}{\textwidth}
        \centering
        \caption{HFC}
        \begin{tabular}{llc|c|c|c}
        \hline
        & & \multicolumn{4}{c}{\textbf{Improvement over}}  \\
        & & \multicolumn{2}{c|}{\textbf{No Explanations}} & \multicolumn{2}{c}{\textbf{Human Explanations}} \\
        \hhline{~~----}
        \textbf{Explainer} & \textbf{Method} & \textbf{Accuracy} & \textbf{Macro F1} & \textbf{Accuracy} & \textbf{Macro F1} \\
        \hline
        \textbf{GPT-4o} & \textbf{Few-shot} & $0.037_{\stddev{0.045}}$ & $0.074_{\stddev{0.093}}$ & $0.010_{\stddev{0.045}}$ & $0.027_{\stddev{0.093}}$ \\
         & \textbf{Zero-shot} & $0.046_{\stddev{0.039}}$ & $0.092_{\stddev{0.074}}$ & $0.019_{\stddev{0.039}}$ & $0.044_{\stddev{0.074}}$ \\
        \hline
        \textbf{Gemma2} & \textbf{Few-shot} & $0.037_{\stddev{0.034}}$ & $0.076_{\stddev{0.072}}$ & $0.010_{\stddev{0.034}}$ & $0.028_{\stddev{0.072}}$ \\
         & \textbf{Zero-shot} & $0.039_{\stddev{0.040}}$ & $0.082_{\stddev{0.072}}$ & $0.011_{\stddev{0.040}}$ & $0.035_{\stddev{0.072}}$ \\
        \hline
        \textbf{Llama3} & \textbf{Few-shot} & $0.070_{\stddev{0.054}}$ & $0.119_{\stddev{0.093}}$ & $0.042_{\stddev{0.054}}$ & $0.071_{\stddev{0.093}}$ \\
         & \textbf{Zero-shot} & $0.060_{\stddev{0.062}}$ & $0.095_{\stddev{0.127}}$ & $0.033_{\stddev{0.062}}$ & $0.047_{\stddev{0.127}}$ \\
        \hline
        \textbf{Mixtral} & \textbf{Few-shot} & $0.037_{\stddev{0.042}}$ & $0.066_{\stddev{0.088}}$ & $0.009_{\stddev{0.042}}$ & $0.019_{\stddev{0.088}}$ \\
         & \textbf{Zero-shot} & $0.056_{\stddev{0.049}}$ & $0.107_{\stddev{0.091}}$ & $0.028_{\stddev{0.049}}$ & $0.059_{\stddev{0.091}}$ \\

         \hline
        \end{tabular}        
    \end{subtable}
\end{table}

\subsection{Influence of Explanations on the Performance of LLMs}
\textbf{LLM explanations struggle to outperform the no-explanation baseline.} As both Table \ref{tab:llms_comparison} and
Figures \ref{fig:esnli_llms_macro}, \ref{fig:hfc_llms_macro}
show, in most cases, providing the classifier LLMs with LLM-generated explanations does not lead to better performance than having no explanations. This is in stark contrast to the results for the PLMs, where having explanations always led to benefits over the baseline. This difference might be because the LLMs used as classifiers are not explicitly trained on the explanations, and thus do not learn to use that information.  
{The logic-based explanations of e-SNLI are akin to the CoT mechanism that LLMs deploy when answering reasoning questions. These explanations only improved the performance of PLMs, which seemingly do not have such a mechanism in their predictive process, but hurt the performance of LLMs, where the explanations clashed with their internal reasoning. Conversely, the summary-style explanations of HealthFC serve to provide additional context and background knowledge, and helped the PLMs and only Llama among LLMs.}
Our findings highlight the importance of tailoring explanation strategies to both the model type and task characteristics.

\begin{table}[t!]
\scriptsize
\small

    \centering
    \caption{\label{tab:llms_comparison}Performance \textbf{impact} by LLM-generated explanations over the baseline of no explanations and human-written explanations, \textbf{averaged over the three LLMs} we have used as classifiers. Subscripts indicate standard deviations.}

    \begin{subtable}{\textwidth}
        \centering
        \caption{e-SNLI}
        \begin{tabular}{llc|c|c|c}
        \hline
        & & \multicolumn{4}{c}{\textbf{Improvement over}}  \\
        & & \multicolumn{2}{c|}{\textbf{No Explanations}} & \multicolumn{2}{c}{\textbf{Human Explanations}} \\
        \hhline{~~----}
        \textbf{Explainer} & \textbf{Method} & \textbf{Accuracy} & \textbf{Macro F1} & \textbf{Accuracy} & \textbf{Macro F1} \\
        \hline
        \textbf{GPT-4o} & \textbf{Few-shot} & $-0.072_{\stddev{0.047}}$ & $-0.078_{\stddev{0.054}}$ & $-0.226_{\stddev{0.047}}$ & $-0.233_{\stddev{0.054}}$ \\
         & \textbf{Zero-shot} & $-0.071_{\stddev{0.061}}$ & $-0.078_{\stddev{0.068}}$ & $-0.226_{\stddev{0.061}}$ & $-0.234_{\stddev{0.068}}$ \\
        \hline
        \textbf{Gemma2} & \textbf{Few-shot} & $-0.137_{\stddev{0.056}}$ & $-0.149_{\stddev{0.066}}$ & $-0.292_{\stddev{0.056}}$ & $-0.304_{\stddev{0.066}}$ \\
         & \textbf{Zero-shot} & $-0.155_{\stddev{0.060}}$ & $-0.169_{\stddev{0.072}}$ & $-0.310_{\stddev{0.060}}$ & $-0.325_{\stddev{0.072}}$ \\
        \hline
        \textbf{Llama3} & \textbf{Few-shot} & $-0.119_{\stddev{0.057}}$ & $-0.130_{\stddev{0.067}}$ & $-0.274_{\stddev{0.057}}$ & $-0.285_{\stddev{0.067}}$ \\
         & \textbf{Zero-shot} & $-0.100_{\stddev{0.062}}$ & $-0.113_{\stddev{0.072}}$ & $-0.255_{\stddev{0.062}}$ & $-0.269_{\stddev{0.072}}$ \\
        \hline
        \textbf{Mixtral} & \textbf{Few-shot} & $-0.106_{\stddev{0.052}}$ & $-0.116_{\stddev{0.058}}$ & $-0.261_{\stddev{0.052}}$ & $-0.271_{\stddev{0.058}}$ \\
         & \textbf{Zero-shot} & $-0.125_{\stddev{0.060}}$ & $-0.135_{\stddev{0.069}}$ & $-0.280_{\stddev{0.060}}$ & $-0.291_{\stddev{0.069}}$ \\

         \hline
        \end{tabular}
    \end{subtable}

    \begin{subtable}{\textwidth}
        \centering
        \caption{HFC}
        \begin{tabular}{llc|c|c|c}
        \hline
        & & \multicolumn{4}{c}{\textbf{Improvement over}}  \\
        & & \multicolumn{2}{c|}{\textbf{No Explanations}} & \multicolumn{2}{c}{\textbf{Human Explanations}} \\
        \hhline{~~----}
        \textbf{Explainer} & \textbf{Method} & \textbf{Accuracy} & \textbf{Macro F1} & \textbf{Accuracy} & \textbf{Macro F1} \\
        \hline
        \textbf{GPT-4o} & \textbf{Few-shot} & $-0.067_{\stddev{0.057}}$ & $-0.045_{\stddev{0.074}}$ & $0.023_{\stddev{0.057}}$ & $0.055_{\stddev{0.074}}$ \\
         & \textbf{Zero-shot} & $-0.076_{\stddev{0.066}}$ & $-0.044_{\stddev{0.068}}$ & $0.014_{\stddev{0.066}}$ & $0.056_{\stddev{0.068}}$ \\
        \hline
        \textbf{Gemma2} & \textbf{Few-shot} & $-0.039_{\stddev{0.066}}$ & $-0.036_{\stddev{0.096}}$ & $0.051_{\stddev{0.066}}$ & $0.063_{\stddev{0.096}}$ \\
         & \textbf{Zero-shot} & $-0.043_{\stddev{0.048}}$ & $-0.026_{\stddev{0.061}}$ & $0.047_{\stddev{0.048}}$ & $0.074_{\stddev{0.061}}$ \\
        \hline
        \textbf{Llama3} & \textbf{Few-shot} & $0.014_{\stddev{0.044}}$ & $0.055_{\stddev{0.053}}$ & $0.104_{\stddev{0.044}}$ & $0.155_{\stddev{0.053}}$ \\
         & \textbf{Zero-shot} & $0.052_{\stddev{0.036}}$ & $0.098_{\stddev{0.042}}$ & $0.142_{\stddev{0.036}}$ & $0.198_{\stddev{0.042}}$ \\
        \hline
        \textbf{Mixtral} & \textbf{Few-shot} & $-0.075_{\stddev{0.061}}$ & $-0.038_{\stddev{0.059}}$ & $0.015_{\stddev{0.061}}$ & $0.061_{\stddev{0.059}}$ \\
         & \textbf{Zero-shot} & $-0.037_{\stddev{0.058}}$ & $0.006_{\stddev{0.057}}$ & $0.053_{\stddev{0.058}}$ & $0.106_{\stddev{0.057}}$ \\

         \hline
        \end{tabular}
    \end{subtable}
\end{table}

\noindent\textbf{LLM explanations come close to human explanations.} Averaged over the classifier LLMs, the results in Table \ref{tab:llms_comparison} show that on e-SNLI human explanations are considerably more beneficial than LLM explanations, with improvements in accuracy around 20-30\%. On the HealthFC dataset, LLM explanations are more helpful, but with smaller differences in accuracy ranging from as low as 1\% to 20\%. These results, combined with the comparisons per model in 
Figures \ref{fig:esnli_llms_macro}, \ref{fig:hfc_llms_macro}
indicate that human explanations are more helpful for LLMs than LLM-generated explanations in more cases and more strongly. Averaged over the classifier LLMs, the results in Table \ref{tab:llms_comparison} show that on e-SNLI human explanations are considerably more beneficial than LLM explanations, with improvements in accuracy around 20-30\%. 

\noindent\textbf{Effect of human explanations on LLMs varies strongly between datasets and models.} Finally,
Figures \ref{fig:esnli_llms_macro}, \ref{fig:hfc_llms_macro}
show that while human explanations consistently lead to improvements over the baseline on e-SNLI, they only improve the performance of Llama 3 on HFC, and to a smaller extent. With both GPT-4o mini and Qwen 2.5, human explanations instead lead to performance decreases of around 10\%. These results again support the claim that LLMs are less successful in using the provided explanations to their benefit compared to PLMs fine-tuned on the explanations, and that the extent to which the LLMs make use of the explanations varies between datasets and LLMs.

\noindent\textbf{LLMs do not necessarily favor their own explanations.} We show in
Figures \ref{fig:esnli_llms_macro}, \ref{fig:hfc_llms_macro}
that particularly comparing GPT-4o mini and Llama3, providing explanations generated by the models from the same model family as the classifier model do not necessarily lead to better performance than providing explanations generated by models from different families. On e-SNLI both models perform best with explanations generated by GPT-4o, and on HFC, with explanations generated by Llama3. This implies that the impact of the explanations rely more on the model generating the explanations rather than whether the explanation and the classifier models belong to the same family.

\subsection{Different Types of Explanations} \label{sec:types}
The explanations in the two datasets serve a different purpose. For e-SNLI, the explanations aim to clarify the \emph{logical reasoning} process using which an entailment label was determined (e.g., \textit{The person is standing, therefore they cannot be sitting}). On the other hand, explanations in the HealthFC dataset serve as a \emph{summary} of the full-text evidence articles and aim to describe what was discovered (e.g., \textit{Analyzed studies have found a positive effect of the drug on the illness}). This could explain the differences between performances of different models for different explanations. The logic-based explanations of e-SNLI are akin to the chain-of-thought (CoT) mechanism that LLMs deploy when answering reasoning questions. These explanations only improved the performance of PLMs, which seemingly do not have such a mechanism in their own predictive process, but hurt the performance of LLMs, where the explanations clashed with their internal reasoning process. Conversely, the summary-style explanations of HealthFC serve to provide additional context and background knowledge to the models, which could explain why they improved the performance of PLMs and, in some cases, even LLMs. Providing additional evidence in prompts to models in an explanatory way augments their knowledge state and leads to improved final reasoning predictions. 

In addition, we also experimented with providing randomly chosen explanations from the datasets but observed worse performance than providing actual explanations. This implies that, unsurprisingly, the content of the explanations influences the models' predictions

\section{Conclusion} In this work, we introduced a novel LLM-based framework for automatically generating textual explanations for NLI tasks. Our evaluation demonstrates that these automated rationales exhibit competitive quality to human annotations and can significantly enhance downstream model performance. This framework presents new opportunities for leveraging LLM explanations to augment non-explainable datasets and improve downstream model classification performance for both PLMs and LLMs. This work in particular highlights the potential of leveraging NLEs to improve LLMs' reasoning performance.

\noindent\textbf{Future work} will explore extending the framework to a broader set of datasets to encompass a wider range of tasks and complexities and further refine prompt engineering and explanation generation via refinement techniques \cite{wang-etal-2025-cross-refine}
, verification and refinement \cite{quan-etal-2024-NLEs-refinement}, and consistency fine-tuning \cite{chen-etal-2025-EC-Fine-tune}. Additionally, incorporating emerging evaluation metrics such as TIGERScore \cite{jiang2024tigerscore} and Prometheus \cite{kim2024prometheus} will enable more comprehensive quality assessments, while comparisons with advanced reasoning LLMs like OpenAI o3 and DeepSeek R1 could further validate our approach.
In addition, we plan to extend our selection of LLMs used for generating explanations by experimenting with LLMs from the same family of different sizes (e.g., Gemma-9b vs Gemma-27b) to measure the impact of size on the quality of explanations per the metrics used in this study. Finally, another point of improvement is measuring and improving the faithfulness of self-explanations by LLMs \cite{parcalabescu2023measuring}, as we have observed that when asked to output the most important words for their predictions, LLMs frequently assign high importance to peripheral words in the prompt such as those describing the labels or denoting parts of the input such as the explanations provided.

\noindent\textbf{Limitations.} Our study is constrained by the sizes of the datasets considered and by the inherent challenges of evaluation metrics (e.g., MAUVE requires large output samples, and API costs for G-Eval can be prohibitive). In addition, the selection of LLMs we employed for generating explanations is limited by using one size per model family, as discussed in the future work, the study could benefit from extending this selection to models from the same family and of different sizes to systematically measure the effect of size on LLMs of same family. 
 Despite these limitations, our findings underscore the strong potential of natural language explanations by LLM from different families and sizes in extending datasets with rationales and improving PLMs and LLMs performance in classification tasks.

\begin{credits}
\subsubsection{\ackname} 
We would like to thank the anonymous reviewers for their helpful suggestions. This research has been supported by the German Federal Ministry of Education and Research (BMBF) grant 01IS23069 Software Campus 3.0 (TU München). 

\end{credits}

%
%
%

\bibliographystyle{splncs04}
\bibliography{icann2025arxiv}

\newpage
\section*{Appendix}

\subsection{Further Datasets Details} \label{app:datasets}

HealthFC is officially a dataset for automated fact-checking (claim verification), it is common to model this task as an NLI task. In this case, the hypothesis is the input claim being fact-checked, and the premise is the evidence text. Since the original evidence articles in HealthFC were very long, we took only the top 5 most relevant evidence sentences selected by the original authors. The fact-checking labels \textit{supported, refuted}, and \textit{not enough information} are then mapped to the NLI labels \textit{entailment, contradiction}, and \textit{neutral}, respectively. Table \ref{tab:dataset_examples} presents an example of an instance from each dataset. 

\begin{table}[!ht]
    \centering
    \caption{Instance example from e-SNLI and HealthFC datasets.}
    \setlength{\tabcolsep}{4pt}
    \renewcommand{\arraystretch}{1.3}
    \small
    \scriptsize

    \begin{tabularx}{\textwidth}{lX X X >{\raggedright\arraybackslash}XX}
        \toprule
        \textbf{Dataset} & \textbf{Premise} & \textbf{Hypothesis} & \textbf{Label} & \textbf{Human Explanation} \\
        \midrule
        e-SNLI & A man leans against a pay phone while reading a paper. & The man is standing and holding a newspaper. & Entailment & If the man is reading a paper, he is reading a newspaper. \\
        \midrule
        HealthFC & Studies show that masks can slow the spread of the coronavirus after three years of the pandemic, as there are now relatively meaningful data supporting this. Surgical masks, in turn, seem to reduce the risk of self-infection with the coronavirus. & Can masks reduce corona infections when worn by a large proportion of the population? & Entailment & International studies suggest that the number of corona infections decreases when many people wear masks, whether fabric, surgical, or FFP2 masks. However, it is not possible to precisely quantify how great the protective effect actually is. It probably depends on the type of masks and the proportion of people who wear them. \\
        \bottomrule
    \end{tabularx}
    \label{tab:dataset_examples}
\end{table}

\subsection{Generating NLEs with LLMs} \label{app:nle-generation}

For the LLMs used for generating NLEs, we used the APIs provided by Groq \footnote{\url{https://groq.com/}} for Mixtral-7B, Gemma2-9B, and LLama3-70B while for GPT-4o mini we use OpenAI APIs \footnote{\url{https://platform.openai.com/}}. 

For the zero-shot setting on e-SNLI, we prompt the LLMs as follows: 
\begin{tcolorbox}[colback=cyan!10,colframe=black!50, sharp corners]
\begin{quote}
Given the following:\\
Premise: \texttt{"\{premise\}"} \\
Hypothesis: \texttt{"\{hypothesis\}"} \\
Label: \texttt{"\{label\_num\}"} (where entailment = 0, neutral = 1, contradiction = 2) \\
Provide exactly one sentence that directly connects the premise to the hypothesis. \\
Do not include any prefixes like "Explanation:" or "Here is the explanation." \\
Start directly with the explanation sentence. \\
The explanation should not explicitly hint at the label or contain the label itself in any form. \\
Focus solely on reasoning that connects the premise to the hypothesis without revealing the classification.
\end{quote}
\end{tcolorbox}

on HeathFC, the prompt is: 
\begin{tcolorbox}[colback=green!10,colframe=black!50, sharp corners]
\begin{quote}
Given the following:
Claim: \texttt{"\{claim\}"} \\
Sentences: \texttt{"\{evidence\}"} \\
Label: \texttt{"\{label\_number\}"} (where Supported = 0, Not enough information = 1, Refuted=2) \\
Now, given the provided claim, sentences, and label, provide only the concise explanation in one sentence, directly referencing the claim and the provided sentences. \\
Do not include any prefixes like "Explanation:" or "Here is the explanation." \\
Start directly with the explanation sentence. \\
The explanation should not explicitly hint at the label or contain the label itself in any form. \\
Focus solely on reasoning that connects the premise to the hypothesis without revealing the classification.
\end{quote}
\end{tcolorbox}

For the few-shot setting, we extend the prompt templates with four (\textit{premise-hypothesis-explanation}) instances from the dataset as examples.

\subsection{Results and Examples of NLE Generation with LLMs} \label{app:nle-generation}

We generated natural language explanations using four LLMs for e-SNLI and HealthFC, under zero-shot and few-shot settings. This results in a total of 16 additional LLM-generated explanation sets (\textit{datasets × LLMs × settings}), which extend the original datasets. We set the temperature to zero during explanation generation to ensure deterministic outputs. Tables \ref{tab:esnli_explanations_examples} \& \ref{tab:hfc_explanations_examples} provides examples of explanations from the four LLMs for one instance in e-SNLI and HealthFC, respectively. 

\begin{table}[t!]
    \caption{Examples of LLM-generated explanations of the four LLMs for zero-shot and few-shot prompts for the
    same instance in e-SNLI presented in Table
    \ref{tab:dataset_examples})
  }
    \centering
    \setlength{\tabcolsep}{4pt}
    \renewcommand{\arraystretch}{1.3}
    \scriptsize
    \begin{tabularx}{\textwidth}{lX X X X}
        \toprule
        & \textbf{Gemma2-9B} & \textbf{Mistral-7B} & \textbf{Llama3-70B} & \textbf{GPT4o-mini} \\
        \midrule
         \textbf{Zero-shot} & 
        The premise states the man is leaning against a pay phone, which implies he is standing, and reading a paper suggests he is holding it. & 
        The man's posture of leaning against the pay phone implies he is standing, and his engagement with the paper suggests he is holding it, possibly reading a newspaper. & 
        The man's posture, with his weight against the pay phone, suggests that he is standing, and the object he is reading is likely a newspaper, given its typical format and size. & 
        The man is indeed standing while he is reading a newspaper, as indicated by his position against the pay phone. \\
        \midrule
         \textbf{Few-shot} & 
        The premise states the man is reading a paper, which could be a newspaper. & 
        The man, who is leaning against a pay phone, is engaging in the activity of reading a paper, which is a type of holding a newspaper. & 
        Since the man is reading a paper, it is likely that the paper is a newspaper. & 
        The man is leaning against a pay phone, which suggests he is standing while he reads the paper. \\
        \bottomrule
    \end{tabularx}
    \label{tab:esnli_explanations_examples}
\end{table}

\begin{table}[!]
    \caption{Examples of LLM-generated explanations for zero-shot and few-shot prompts for 
    the same instance in HealthFC presented in Table \ref{tab:dataset_examples})
    }
    \centering
    \setlength{\tabcolsep}{4pt}
    \renewcommand{\arraystretch}{1.3}
    \scriptsize

    \begin{tabularx}{\textwidth}{lXXXX}
        \toprule
        & \textbf{Gemma2-9B} & \textbf{Mistral-7B} & \textbf{Llama3-70B} & \textbf{GPT4o-mini} \\
        \midrule
        \textbf{ZS} & 
        The sentences suggest that masks can slow the spread of the corona virus and reduce the risk of self-infection, implying a potential for reducing infections. & 
        The first sentence supports the claim by indicating that studies show masks can slow down the spread of the Corona virus, while the second sentence provides additional evidence that surgical masks specifically can reduce the risk of self-infection. & 
        The claim that masks can reduce corona infections when worn by a large proportion of the population is supported by the evidence that masks can slow down the spread of the Corona virus and reduce the risk of infecting oneself with the virus. & 
        The sentences indicate that studies have shown masks can slow the spread of the coronavirus and that surgical masks can reduce the risk of self-infection, suggesting that masks worn by a large proportion of the population could indeed help reduce infections. \\
        \midrule
        \textbf{FS} & 
        The provided sentences suggest that masks can slow the spread of the corona virus, but they do not explicitly state whether this effect is significant enough when a large proportion of the population wears them. & 
        The first sentence provides evidence that masks can slow down the spread of the Corona virus, supporting the claim that masks can reduce corona infections when worn by a large proportion of the population. & 
        The claim that masks can reduce corona infections when worn by a large proportion of the population is supported by studies showing that masks can slow down the spread of the Corona virus and reduce the risk of infection. & 
        Studies indicate that masks can slow the spread of the coronavirus, suggesting that widespread mask usage may help reduce infections in the population. \\
        \bottomrule
    \end{tabularx}

    \label{tab:hfc_explanations_examples}
\end{table}

\subsection{Evaluating NLEs} \label{app:evaluation}

We evaluate the quality of the LLM-generated explanations using BLEU, ROUGE, BERTScore, MAUVE, and G-Eval, as described in Section~\ref{sec:evaluation_metrics}.
We compute G-Eval using the DeepEval library\footnote{\url{https://github.com/confident-ai/deepeval}}, while the remaining metrics are implemented via rouge-score\footnote{\url{https://pypi.org/project/rouge-score/}}, BLEU\footnote{\url{https://www.nltk.org/_modules/nltk/translate/bleu_score.html}}, BERTScore\footnote{\url{https://pypi.org/project/bert-score/}}, and MAUVE\footnote{\url{https://github.com/krishnap25/mauve}} libraries. 

\textbf{For computing G-Eval scores in our pipeline}, we use GPT-3.5-turbo as the judge model to limit potential bias as GPT-4o mini is among the LLMs used to generate explanations. We apply the following prompt template to compute the scores: 

\begin{tcolorbox}[colback=red!10,colframe=black!50, sharp corners]
Evaluation Criteria:\\
Human-Likeness - Measures how natural and human-like an LLM's explanations appear, assigning a score based on similarity to human writing.\\
Evaluation Steps: \\
1. Analyze the human explanation for style, clarity, and focus.\\
2. Examine the LLM-generated explanation, noting patterns and structure.\\
3. Compare both explanations in terms of writing style, coherence, and emphasis.\\
4. Identify LLM-specific artifacts (e.g., excessive verbosity, over-explanation).\\
5. Assign a score from 1 (least human-like) to 5 (most human-like) based on these comparisons.
\normalfont
\end{tcolorbox}

\subsection{Experimenting with LLMs for NLI} \label{app:models}

We adopt a zero-shot inference approach without fine-tuning. Instead, the generated explanations are directly appended to the hypothesis in the prompt. This strategy is driven by practical considerations: fine-tuning large-scale LLMs would incur substantial computational costs, require specialized hardware, and is often infeasible given the models’ enormous parameter sizes. We prompt the LLMs using the following template, with the optional explanation:

\begin{tcolorbox}[colback=gray!10,colframe=black!50, sharp corners]
   Classify the relationship as one of 'entailment' (0), 'neutral' (1), or 'contradiction' (2).\\
    Premise: \textit{<premise>} \\ 
    Hypothesis: \textit{<hypothesis>} \\
    Explanation: \textit{<explanation>} \\
    Use the explanation provided and choose only one label number and the six most important keywords used to answer. Example: 0; word1, ..., word6.
\normalfont
\end{tcolorbox}

\end{document}